\documentclass[letterpaper, 10 pt, conference, hidelinks]{ieeeconf}
\IEEEoverridecommandlockouts                              
\usepackage{graphicx,dblfloatfix}
\usepackage{amsmath} 
\usepackage{amssymb}  
\usepackage{subcaption}
\usepackage{array}
\usepackage{cite}   
\usepackage{hyperref}
\usepackage{float}
\usepackage{mathtools}
\usepackage{multirow}
\usepackage{siunitx}

\usepackage{rotating,booktabs}

\usepackage{microtype}
\usepackage{gensymb}

\hypersetup{
    colorlinks=false,
    linkcolor=black,
    urlcolor=black,
}

\def\showcomments{1}
\usepackage{xcolor}
\newcommand{\peter}[1]{{
    \if\showcomments1
        \color{red}PIC: #1
    \fi
}}

\newcommand{\juxicomment}[1]{{
    \if\showcomments1
        \color{red}JL: #1
    \fi
}}

\title{\LARGE \bf Reactive Base Control for On-The-Move Mobile Manipulation in Dynamic Environments}

\author{Ben Burgess-Limerick$^{1, 2}$, Jesse Haviland$^{1, 2}$, Chris Lehnert$^{1}$,  Peter Corke$^{1}$
\thanks{This research was supported by the QUT Centre for Robotics and CSIRO Data61.}
\thanks{$^{1}$ Queensland University of Technology Centre for Robotics (QCR), Brisbane, Australia }
\thanks{$^{2}$ CSIRO Data61, Brisbane, Australia }
}
\begin{document}
\maketitle
\thispagestyle{empty}
\pagestyle{empty}
\begin{abstract}
We present a reactive base control method that enables high performance mobile manipulation on-the-move in environments with static and dynamic obstacles. Performing manipulation tasks while the mobile base remains in motion can significantly decrease the time required to perform multi-step tasks, as well as improve the gracefulness of the robot's motion. Existing approaches to manipulation on-the-move either ignore the obstacle avoidance problem or rely on the execution of planned trajectories, which is not suitable in environments with dynamic objects and obstacles. The presented controller addresses both of these deficiencies and demonstrates robust performance of pick-and-place tasks in dynamic environments. The performance is evaluated on several simulated and real-world tasks. On a real-world task with static obstacles, we outperform an existing method by 48\% in terms of total task time. Further, we present real-world examples of our robot performing manipulation tasks on-the-move while avoiding a second autonomous robot in the workspace. See \href{https://benburgesslimerick.github.io/MotM-BaseControl}{benburgesslimerick.github.io/MotM-BaseControl} for supplementary materials.

\end{abstract}

\setcounter{footnote}{2}
\section{Introduction}

Performing mobile manipulation tasks while a robot's base remains in motion can significantly reduce execution time compared with approaches in which the robot pauses to perform manipulations. This is particularly valuable in multi-step tasks where, for example, the robot can be driving towards a second location while picking up an object. Recent works have explored control methods for achieving such Manipulation on-the-Move (MotM) \cite{MotM, ThakarTimeOptimal}. 

Systems that rely on planning are able to claim optimality of generated trajectories and avoid  apriori-known obstacles in the scene. However, planning approaches suffer in real-world environments where they cannot react to perception errors, localisation error, or inaccurate control and are unable to perform in environments with unpredictably moving objects and obstacles. 

In our previous work \cite{MotM}, we introduced an architecture for achieving reactive MotM and demonstrated grasping of unpredictable, dynamic objects while on the move. However, the implementation presented previously uses a mobile base controller that does not consider obstacle avoidance. In this work, we develop a mobile base controller for integration into the reactive MotM architecture that minimises task time while avoiding static and dynamic obstacles. In addition, we augment the quadratic program solved in the redundancy resolution module with a constraint that provides obstacle avoidance for the arm. 

\begin{figure}[t]
\begin{subfigure}{\linewidth}
\centering
\includegraphics[width=\linewidth]{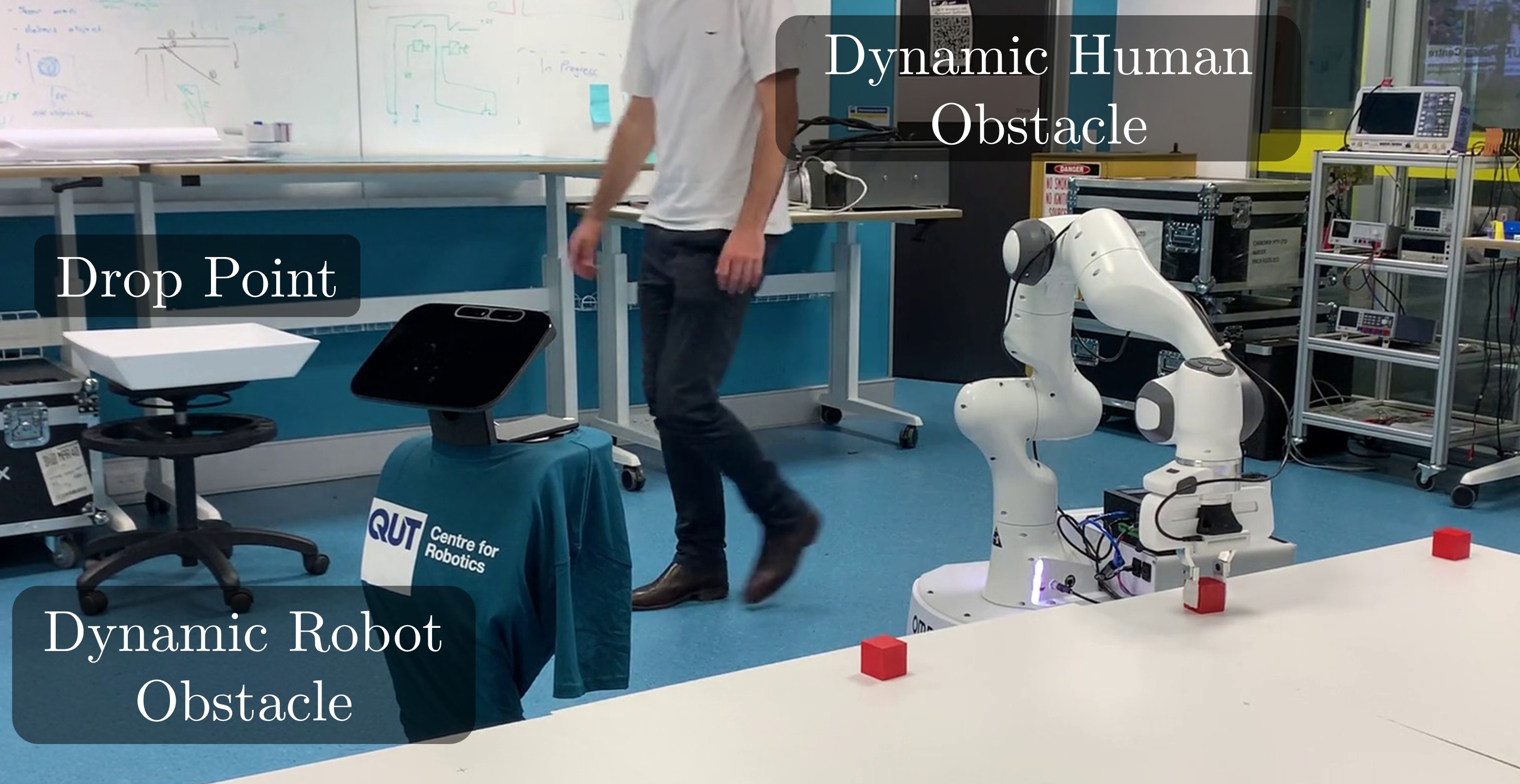}
\caption{Real-world manipulation on-the-move with dynamic obstacles.}
\label{fig:TemiIntReal}
\end{subfigure}

\begin{subfigure}{\linewidth}
\vspace{2mm}
\centering
\includegraphics[width=\linewidth]{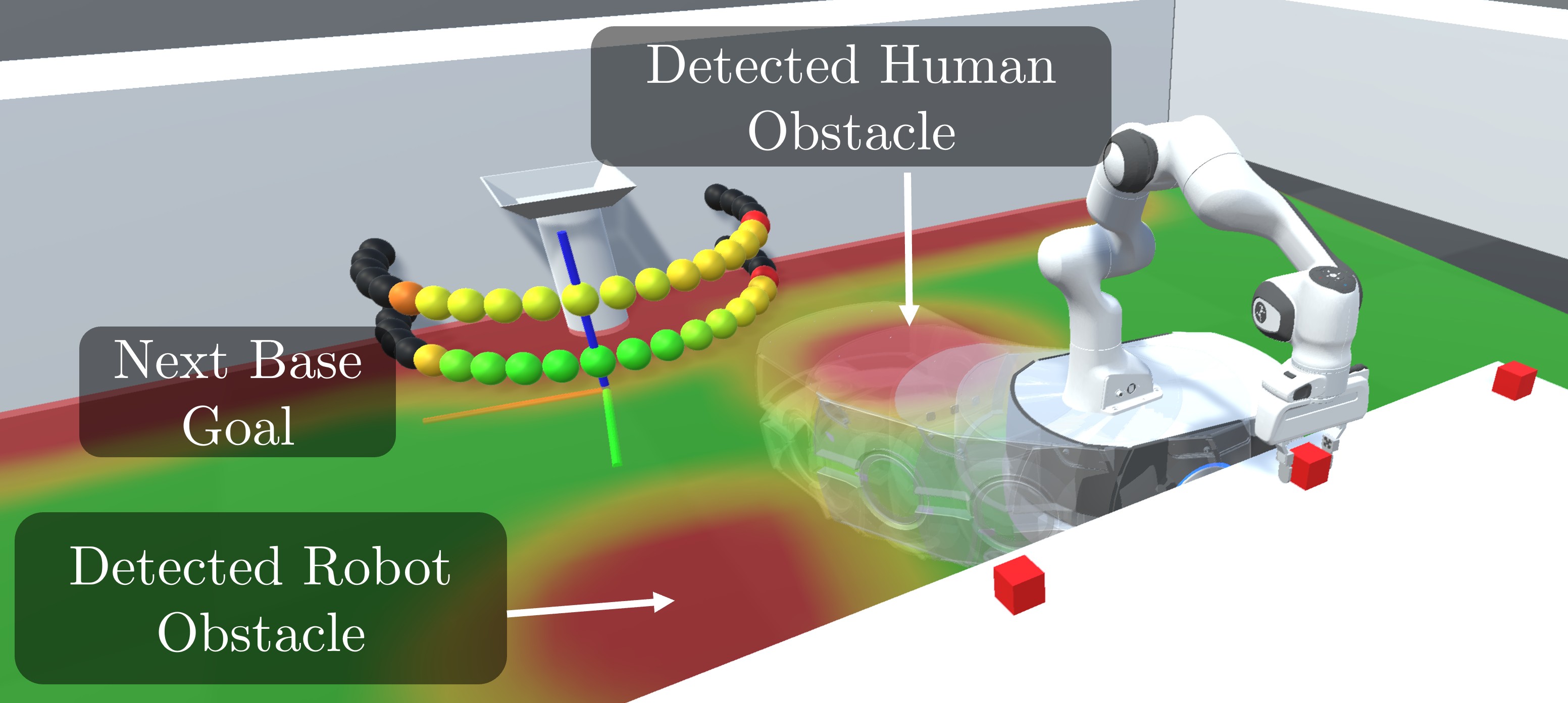}
\caption{Digital Twin rendering of perceived state for Fig. \ref{fig:TemiIntReal}. The system's local plan is illustrated with transparent robots.}
\label{fig:TemiIntDT}
\end{subfigure}
\caption{Our method can perform robust, reactive manipulation on-the-move while avoiding static and dynamic obstacles.}
\vspace{-0.5cm}
\label{fig:TemiInt}
\end{figure}

Our reactive approach enables robust performance in complex environments with dynamic obstacles. Fig. \ref{fig:TemiIntReal} shows a frame from a real-world trial where our system is performing a grasp while on-the-move and avoiding a second robot moving in the scene. The digital twin illustrated in Fig. \ref{fig:TemiIntDT} shows the system's understanding of the space, including the real-time detections of the obstacles in the environment. 

The principal contributions of this work are:
\begin{enumerate}
    \item a reactive base control system that minimises total task time for multi-step tasks by performing manipulation tasks on-the-move
    \item a redundancy resolution module that enables obstacle avoidance for the arm and base while performing manipulation on-the-move tasks
    \item the first demonstration of reactive manipulation on-the-move in real-world environments with static and dynamic obstacles
\end{enumerate}
These capabilities are demonstrated in numerous simulated and real-world experiments.

\section{Related Works}

Base control methods for mobile manipulation guide the mobile base to a pose from which a manipulation target can be reached without violating kinematic constraints or colliding with obstacles. This is often achieved through explicit base control where an optimal base pose is determined and the robot is navigated to that pose with a conventional mobile robot controller \cite{Reister}. Alternatively, implicit or holistic base control methods typically start with a desired end effector motion and use the combined kinematics of the base and manipulator to achieve the motion. 

We review common methods for base control in mobile manipulation and the additional challenges for base control while performing manipulation on-the-move. Recent surveys of control strategies for mobile manipulation include \cite{SandakalumReview}.

\subsection{Explicit Base Control}
\subsubsection{Optimal Base Placement}
Numerous planners have been proposed to compute the base pose from which to best perform a manipulation task. In general, the goal is to find a pose for which the target is reachable with a collision-free configuration of the robot \cite{SerajiBasePlacement}. Approaches typically aim to generate solutions that are optimal against some other metric such as manipulability \cite{VahrenkampRobotPlacement, Jauhri}, or stiffness of the robot \cite{Fan}. Other approaches aim to minimise task time by calculating base poses from which multiple targets can be reached without repositioning \cite{Paus, Xu}. Time efficiency can be optimised on multi-step tasks by choosing a pose based on where the robot must go after the immediate target \cite{Reister, Harada}. Reactivity can be achieved by frequently recomputing the optimal base pose \cite{Reister}.

\subsubsection{Mobile Base Control}
Once an optimal base pose has been computed, the robot is driven to the pose with a mobile base controller. Typically, a hierarchical planner is used that combines a global and local planner to enable reactive obstacle avoidance while driving to the goal pose \cite{Pittner}.

A reactive, Short Term Aborting A* (STAA*) method is presented in \cite{STAA} that demonstrates improved performance in environments with static and dynamic obstacles compared to commonly used local planners. STAA* plans a collision-free global path by searching through a visibility graph, and then computes the intersection of the global path and the border of a local planning region to develop an intermediate goal. A discretised acceleration space is used to sample new states for exploration in a time-bounded A* search. The search uses an obstacle-aware heuristic that ensures exploration along a collision-free path and avoids local minima.

In most cases, mobile manipulators using explicit base controllers consider the base and manipulator motion entirely separately and the arm does not start moving until the base has achieved the desired pose. More recent methods have improved task time by coordinating the arm motion with the base such that the hand arrives at the target at the same time as the base arrives at the desired pose \cite{Reister}. 

\subsection{Holistic Control}
Rather than considering the base and arm motion separately, some approaches combine the subsystems with a holistic controller \cite{HavilandHolistic}. These methods use the combined degrees of freedom from the mobile base and manipulator to achieve a desired end-effector motion. Holistic control of the robot enables reduced task time by moving both components together, as well as an improved ability to optimise secondary objectives such as manipulability \cite{HavilandHolistic}, obstacle avoidance \cite{Shao}, or visibility \cite{HeVisibility} through exploitation of additional degrees of freedom. 

The input to the holistic control system can be from a motion planner and executed under open-loop control \cite{Shao}, or reactive, where the controller uses visual feedback for closed-loop control \cite{HavilandHolistic, Arora}. Model Predictive Control formulations of the problem have been used to enable collision avoidance for both arm and base \cite{Logothetis, Spahn}. A learned base controller is presented in \cite{Honerkamp} which translates a desired end-effector velocity and local occupancy map to base motions that ensure the omnidirectional robot avoids obstacles.

These works demonstrate holistic control of a mobile manipulator and can avoid obstacles. However, they focus only on execution of the immediate goal and do not consider the time efficiency for multi-step tasks achieved by performing a manipulation task while on the move toward the next target. 

\subsection{Manipulation On-The-Move}

The first MotM approaches restricted base motion to constant speed, straight-line motion \cite{ShanMotionPlanning}. Recent works plan collision-free trajectories in cluttered environments that can complete mobile manipulation tasks in minimum time \cite{ThakarTimeOptimal, ThakarManipulatorMotionPlanning, Colombo, Zimmermann, XuOptimizationMotionPlanner}. However, these approaches execute the planned trajectory open-loop and cannot react to dynamic changes in the scene, or compensate for perception and control errors. Consequently, these systems are often failure prone in real-world environments. 

In this work, we present a reactive base control method that avoids obstacles while performing manipulation tasks on-the-move. Further, we enable collision avoidance for the manipulator by implementing the approach described in \cite{HavilandNEO}.

\section{Base Controller}

We introduce several modifications to the global and local planners described in STAA* \cite{STAA} that improve performance for manipulation on-the-move scenarios. 

\subsection{Goal Orientation} The most important addition to STAA* is the inclusion of an orientation to the goal state. Where STAA* considers driving to a point only, we include orientation which enables poses to be achieved that smoothly connects the immediate target with the next goal. 

\begin{figure}[t]
\centerline{\includegraphics[width=\linewidth]{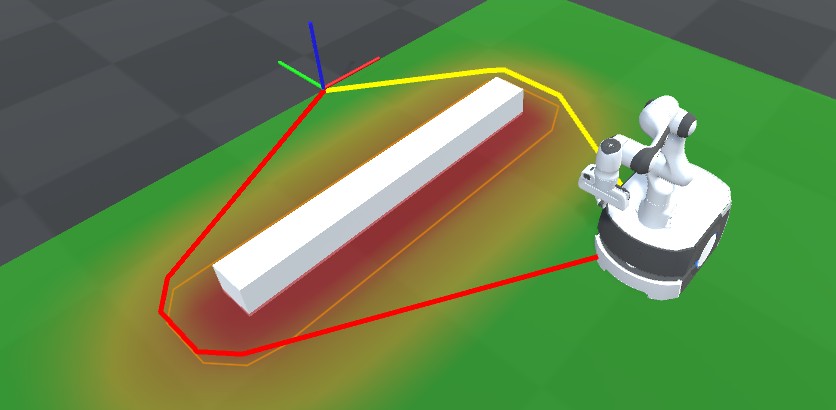}}
\caption{A comparison of global paths generated by our modified STAA* (path in red) and the original (shown in yellow) for an example goal pose. The red $x$ axis of the target frame represents the desired forward direction.}
\label{fig:GlobalPath}
\vspace{-15pt}
\end{figure}

\subsection{Rotation in Global Planner Cost} The addition of orientation to the goal state requires modification to the node cost computation used in the global A* search. STAA* uses only the cumulative distance between nodes along a path to compute the cost. Instead, we consider the PathRTR metric that estimates the time required to both translate and rotate between nodes along a path. Further explanation of the PathRTR metric is presented in \cite{STAA} where it is used for the local planner. Fig. \ref{fig:GlobalPath} illustrates the value of including rotation costs in the global path planner. For the scenario shown, the path in red generated by our modified version encourages the robot to drive a smooth curve around the obstacle connecting the start and end pose. Without rotation costs, the shortest path passes the obstacle on the opposite side and requires significantly more turning. 

\subsection{Search Termination Conditions} The implementation of STAA* presented in \cite{STAA} terminates its search when an explored node is sufficiently close to the goal. To perform manipulation on-the-move, we want to encourage the robot to drive through the goal at high speed. Therefore, we also terminate when the path passes sufficiently close to the target.

\subsection{Reduction of Proximity Grid Penalty} STAA* includes a cost on visiting nodes based on their proximity to obstacles using an inflated occupancy grid. However, to complete mobile manipulation tasks such as picking and placing objects from a table, the robot must necessarily travel close to the table while interacting with objects on it. For example, in Fig. \ref{fig:GlobalPath}, the occupancy grid is represented by the colour of the ground around the robot, with green representing free space, and red representing occupied space. When the target pose for the base is near an obstacle, as is the case when grasping from a table, the penalty applied to nodes near obstacles inhibits the exploration of states near the goal. To limit this effect, we reduce the weight of the occupancy grid cost based on proximity to the object pick or place location. We scale the grid penalty by $k = \max{(0.1, \min{(t_h/3, 1)})}$ where $t_h$ is the estimated time until the goal is achieved.

\subsection{B\'ezier Heuristic} 
\label{Bezier}

\begin{figure}[t]
\centerline{\includegraphics[width=\linewidth]{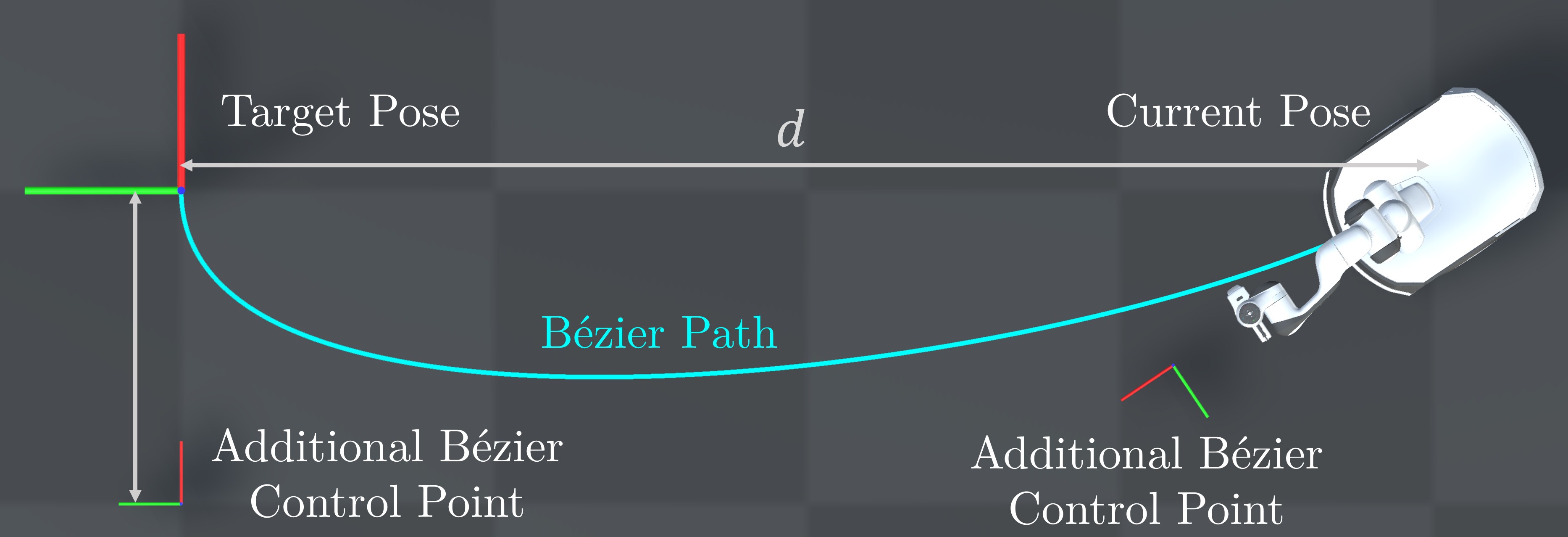}}
\caption{Illustration of the B\'ezier path evaluation. The PathRTR evaluation between the two presented poses consists of Rotate(30\degree), Translate(4\si{m}), Rotate(90\degree). By comparison, the B\'ezier curve shown has a total length of 4.35\si{m}. Assuming the robot has maximum linear and angular velocities of 0.5\si{ms^{-1}} and 100\degree\si{s^{-1}}, the estimated sequential PathRTR cost is 9.2\si{s}, while travelling the B\'ezier path and combining rotation and translation requires only 8.7\si{s}.}
\label{fig:Bezier}
\vspace{-15pt}
\end{figure}

The PathRTR heuristic used for the A* graph search in the local planner assumes that translation and rotation of the robot will be performed at maximum speed, but sequentially. This has a tendency to overestimate costs for states where the optimal path to the goal is a smooth arc of simultaneous rotation and translation. We introduce an additional heuristic based on a B\'ezier curve that is used in place of PathRTR only when the B\'ezier path results in a lower cost. This encourages the exploration of states that can be connected to the goal through smooth curves. The B\'ezier curve is constructed by adding two control points alongside the start and end points given by the current pose and target. The first control point is positioned in front of the robot's current forward direction, and the second is positioned an equal distance behind the desired end pose. The offset distance is chosen to be 25\% of the distance between the current and target pose. The optimal offset distance is a complicated function of the relative target pose and the ratio of the maximum linear and angular velocity capabilities of the robot. However, we find that a value of 25\% results in curves that work well for our robot in practice. Fig. \ref{fig:Bezier} illustrates an example B\'ezier path.

\section{Base Placement}

\label{BasePlacement}

\begin{figure}[t]
\begin{subfigure}{0.49\linewidth}
\centering
\includegraphics[width=\linewidth]{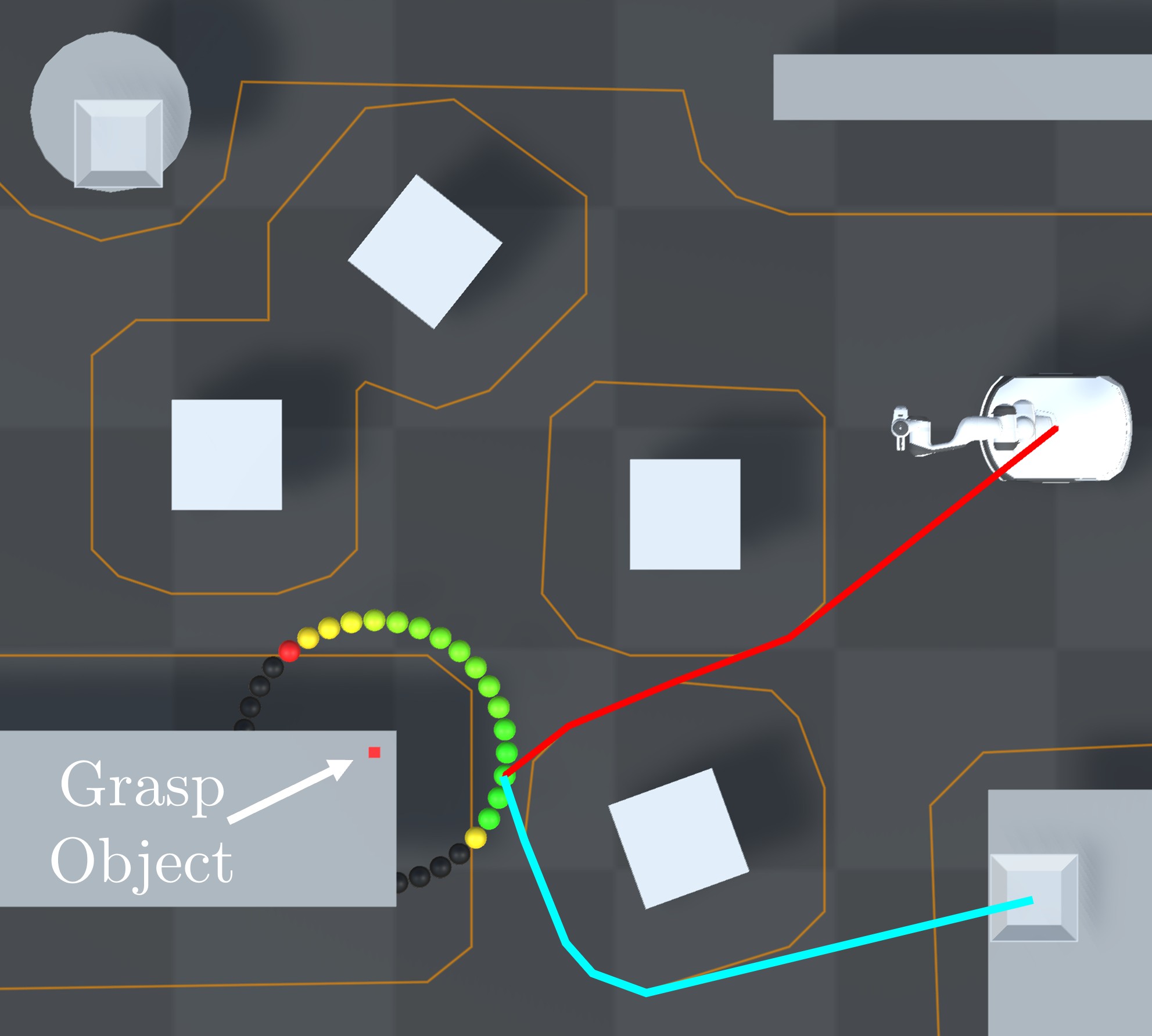}
\caption{Drop Location 1}
\label{fig:BP1}
\end{subfigure}
\begin{subfigure}{0.49\linewidth}
\centering
\includegraphics[width=\linewidth]{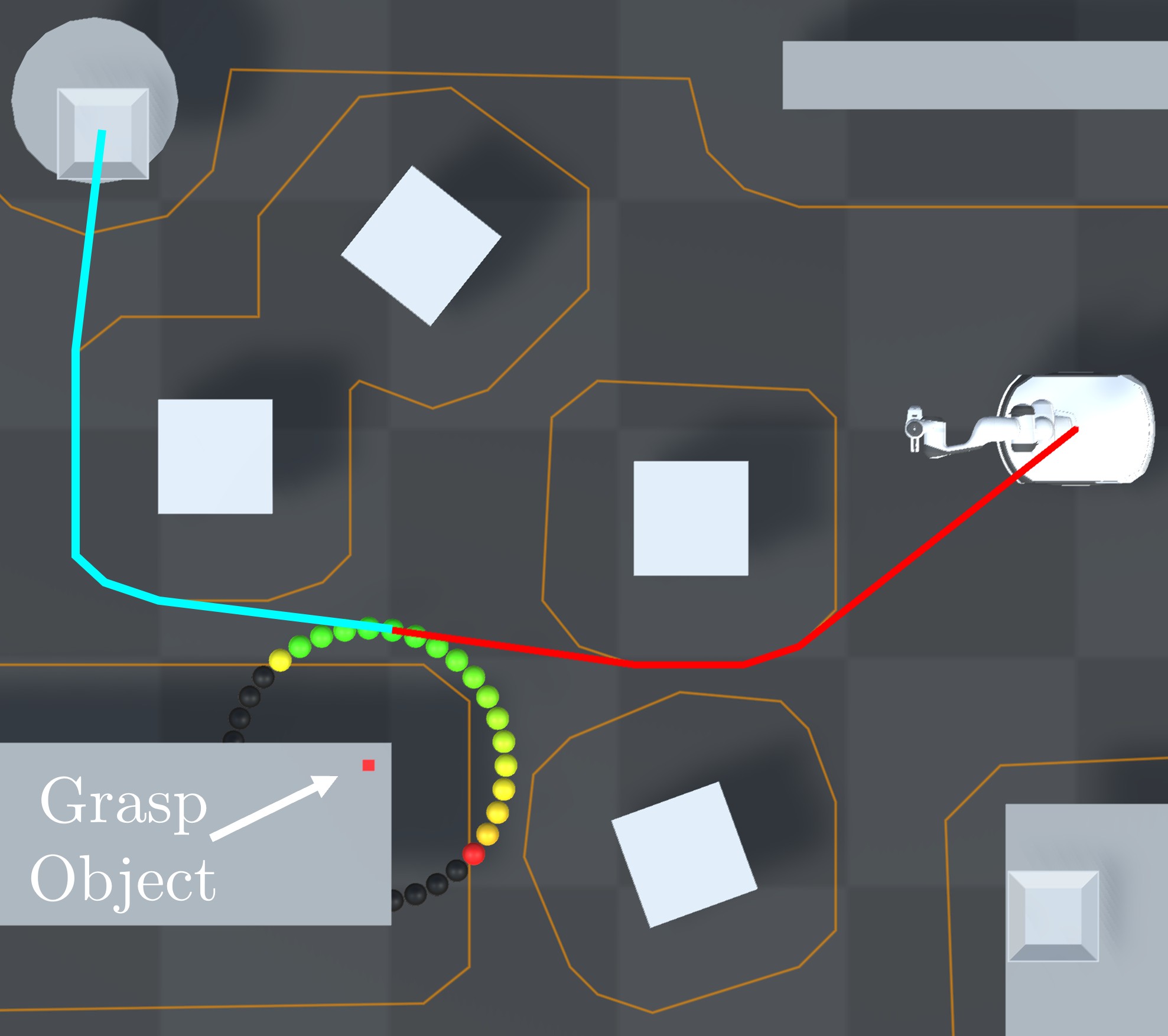}
\caption{Drop Location 2}
\label{fig:BP2}
\end{subfigure}
\caption{Examples of optimal base pose selection for two different drop locations. The red and cyan lines illustrate the path to the candidate and path from candidate to next target respectively for the optimal base placement.}
\vspace{-0.7cm}
\label{fig:BasePlacement}
\end{figure}

The optimal base placement is selected from a discretised set by evaluating the path cost from the current robot pose to candidate base placements as well as from the candidates to the next target (Fig. \ref{fig:BasePlacement}). Candidates are evenly spaced around the target object in 10\degree \ increments for a total of 36 possible base positions. Each position is assigned two possible orientations: the robot's forward vector can be tangential to the circle facing either clockwise or counter-clockwise which gives a total of 72 candidates. 

The radius of the circle on which the candidates are placed is dynamically adjusted between 0.6 \si{m} and 0.8 \si{m}. When no collision-free candidates are available close to the object the radius is increased until a solution is found. The radius limits are defined by the geometry of our robot, 0.6 \si{m} is calculated from the radius of the robot base plus a safety margin, and 0.8 \si{m} is the largest distance at which the robot can still perform manipulation tasks. When no viable candidates are found within these limits, the robot will drive toward the collision-free position closest to the target with the assumption that the system may identify a valid candidate as the map updates with more recent lidar data. 

The path cost for each candidate is evaluated using the PathRTR metric described in \cite{STAA}. The total path cost for the $i$-th candidate is given by the weighted sum of two components
\[C_{i} = C_{i, \text{C}} + 1.05 \cdot C_{i, \text{N}} \]
where $C_{i, \text{C}}$ is the estimated cost from robot to candidate and $C_{i, \text{N}}$ is the estimated cost from the candidate to the next target. The weighting that biases towards minimising $C_{i, \text{N}}$ ensures that the robot continues to drive toward the next target even when close to the object. Decreasing this weight below 1 will tend towards a greedy solution that optimises for the immediate task only without considering the next task. Increasing the weight further will encourage the robot to sacrifice time efficiency on the immediate task in favour of minimising the expected travel time for the next task.

The candidate with lowest $C_i$ is used as the goal for navigation. The method selects base placements that are within manipulation range of the target and efficiently connect the current robot pose with the next target. For example, the spheres in Fig. \ref{fig:BasePlacement} are coloured based on their path cost, with brighter green representing lower cost. The paths for the optimal candidate are also shown.

The path costs for each candidate are reevaluated at each controller step (20 \si{Hz}) to enable reactive control and response to changes in the environment.

\section{Arm Obstacle Avoidance}
\label{ArmAvoidance}
Obstacle avoidance for the manipulator is enabled by modifying the redundancy resolution controller described in \cite{MotM}. This controller solves a Quadratic Program (QP) to calculate joint velocities for a given desired end-effector and base velocity. Using a similar controller, reactive obstacle avoidance for a mobile manipulator is demonstrated in \cite{HavilandNEO}. The controller allows for slack in the achieved end-effector velocity which can be exploited along with redundant degrees of freedom to avoid obstacles. 

Obstacle avoidance is implemented in the QP through the addition of an inequality constraint which limits the velocity of points on the arm when they are close to an obstacle. Further details on the implementation are available in \cite{HavilandNEO}. 

Accurately modelling the environment in sufficient detail for robust collision avoidance in 3D is a difficult problem and outside the scope of this work. In \cite{HavilandNEO}, real-world trials are performed with simulated objects whose pose can be directly observed. In our system, we use the 2D lidar in the robot's base to construct an obstacle map for the arm. Any obstacles detected by the lidar are assumed to be tall enough that the arm should avoid them. When the obstacles are short this is a conservative assumption and the arm will unnecessarily avoid the space above the obstacle. However, when an obstacle is larger above the plane of the lidar (for example a table supported by a central post), the system cannot observe the geometry at the height of the arm and there is the risk of a collision. We mitigate this by providing the system with a pre-generated map of most of the collision geometry in the environment. Additional obstacles detected by the lidar are added to the map. In future work, more detailed collision geometry could be modelled online with a 3D lidar or depth camera, enabling improved obstacle avoidance. 

We query the constructed manipulator obstacle map with a single point in the centre of the robot's gripper rather than computing the closest point on link collision meshes as in \cite{HavilandNEO}. This simplifies the implementation and we find that it achieves acceptable performance in our real-world testing. The end-effector speed toward the object is limited to 
\[
\dot{d_{ro}} \leq \xi \frac{d - d_s}{d_i - d_s}
\]
where $\dot{d_{ro}}$ is the distance between the end-effector and the closest obstacle, $\xi = 0.6$ is a gain controlling the aggressiveness of the obstacle avoidance, $d_s = 0.25$ \si{m} is the minimum distance allowed between end-effector and obstacle, and $d_i = 0.6$ \si{m} is the threshold at which the limit is enabled. For $d_{ro} > d_i$ the constraint is removed from the QP. 

\section{Experiments}
Real-world experiments are performed with our Frankie\footnote{\url{https://github.com/qcr/frankie_docs}} mobile manipulator that consists of an Omron LD-60 differential-drive mobile base and a Franka-Emika Panda 7-DOF manipulator. Robot control is implemented using the Robotics Toolbox \cite{rtb}. The simulation and digital twin environments are implemented in Unity. 

\subsection{Baseline Comparisons}
The method presented in \cite{Reister} optimises task time by selecting base poses that are conditioned on the next target location in a higher-level task. This is similar to the method described in \ref{BasePlacement}, but does not consider the challenges of performing the tasks on-the-move. To meaningfully compare with the work presented in \cite{Reister} we recreate their experiments as closely as possible in simulation and the real-world. \cite{Reister} provides performance data for two baselines, as well as two versions of their proposed approach. These methods are described briefly here:

\begin{enumerate}
    \item \textbf{Fixed Set Baseline:} The robot chooses a base placement from a set of 7 fixed candidates spaced around and facing the object table. Selection amongst the candidates is based on maximising manipulability.
    \item \textbf{Inverse Reachability Maps (IRM) Baseline:} The optimal base placement is selected based on evaluation of a manipulability metric only and does not consider navigation costs. 
    \item\textbf{Greedy \cite{Reister}:} The base placement includes both manipulability and navigation costs associated with a candidate but does not include navigation to the next location. 
    \item\textbf{Sequential \cite{Reister}:} Optimal base placement is determined based on a weighted cost combining manipulability, navigation costs to a candidate, and navigation costs from the candidate to the next location. 
\end{enumerate}
Further details on these methods are available in \cite{Reister}. All data relating to these approaches has been reproduced from \cite{Reister}.

\subsection{Experiment 1: Static Obstacles}
\begin{figure}[t]
\centerline{\includegraphics[width=\linewidth]{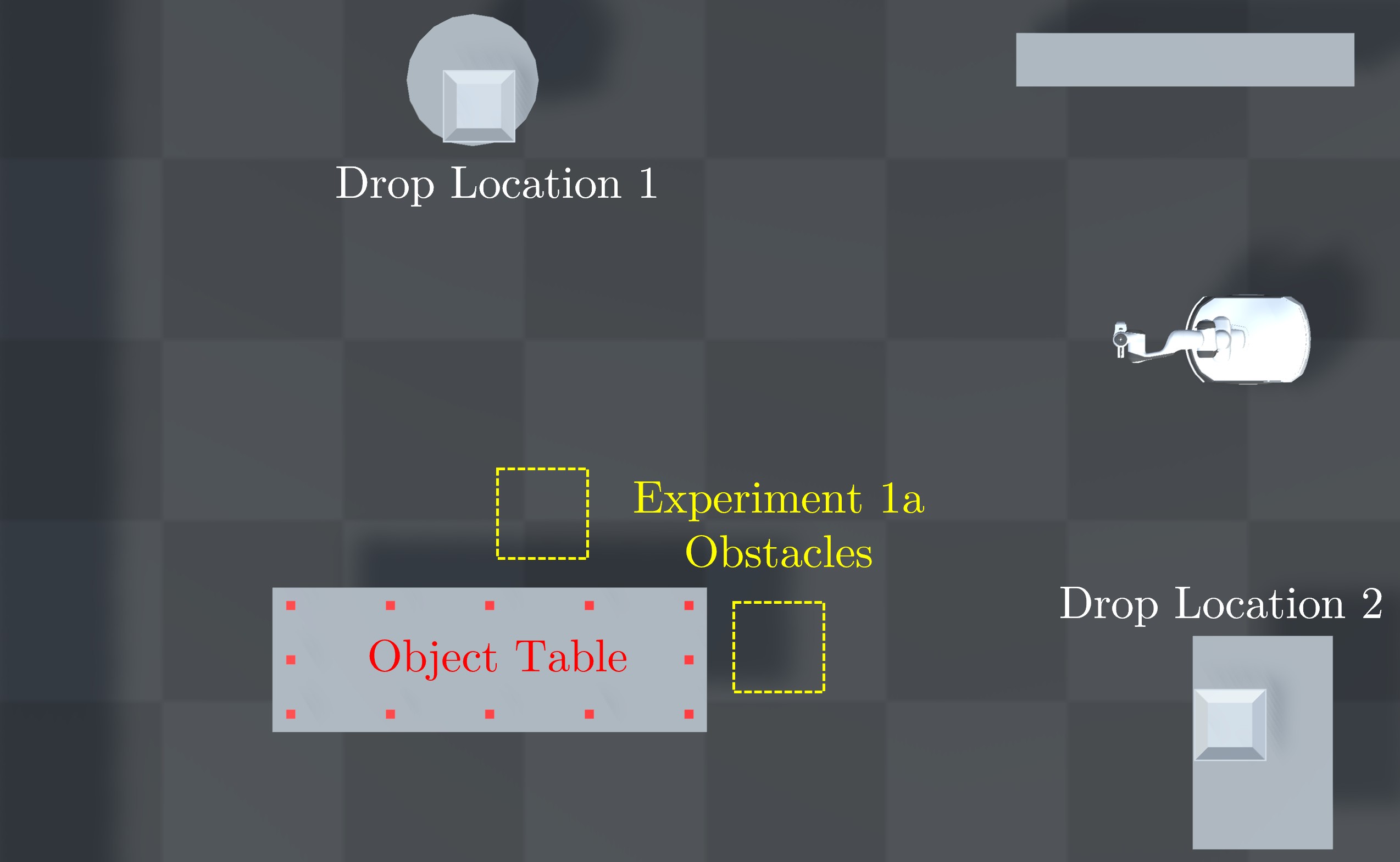}}
\caption{Layout of objects and drop locations for experiment 1. For each trial, 6 of the 12 object positions were randomly selected and an order was generated for them to be alternately delivered to the two drop locations. The yellow squares indicate additional static obstacles for experiment 1a.}                  
\label{fig:Experiment1}
\vspace{-15pt}
\end{figure}

A collection of 6 objects were placed on a 2.4 $\times$ 0.8 \si{m} table and must be transported to two drop locations. Fig. \ref{fig:Experiment1} shows the experiment layout -- a dimensioned diagram is available on our project website. The table and drop points locations are consistent with those used in \cite{Reister}, allowing for meaningful comparison. 

The 6 objects were placed randomly on the 12 possible object locations shown in Fig. \ref{fig:Experiment1} and a random order was assigned. The objects must be picked up in order and transported to alternating drop locations. This is to ensure fair comparison with the experiments in \cite{Reister}, however it should be noted that performance improvements are possible by optimally selecting the order and drop location. 

Our experiment differs from \cite{Reister} in two respects. First, we use red 40 \si{mm} cubic objects to simplify perception and grasp synthesis which is not the focus of this work. Second, we limit object positions to a set of 12 candidates around the perimeter of the table. Our robot has an arm reach of 0.855 \si{m}, significantly less than the 1.55 \si{m} of the ARMAR-6 robot used in \cite{Reister}. Positioning the objects closer to the table's edge allows more room for our robot to perform reactive grasping without colliding with the table. The reach of ARMAR-6 is sufficient to grasp objects on the far side of the table, whereas our robot must drive around the table, increasing the distance to be travelled. We limit the positions to a set of 12 candidates to improve reproducibility of the experiment.

We conducted 50 trials with random object arrangements in simulation and performed 10 trials in the real world. The first real-world object arrangement is hand-crafted for best comparison with the real-world trial presented in \cite{Reister}, and the remaining 9 were randomly chosen. Details of the 10 real-world scenarios are available on our project website. 

\subsection{Experiment 1a: Additional Static Obstacles}
The 50 simulated experiments conducted in Experiment 1 were repeated with the addition of 2 cuboid obstacles (shown in yellow on Fig. \ref{fig:Experiment1}). These objects were positioned to be consistent with those included in scenario 2 presented in \cite{Reister}.

\subsection{Experiment 2: Dynamic Obstacles}
\begin{figure}[t]
\centerline{\includegraphics[width=\linewidth]{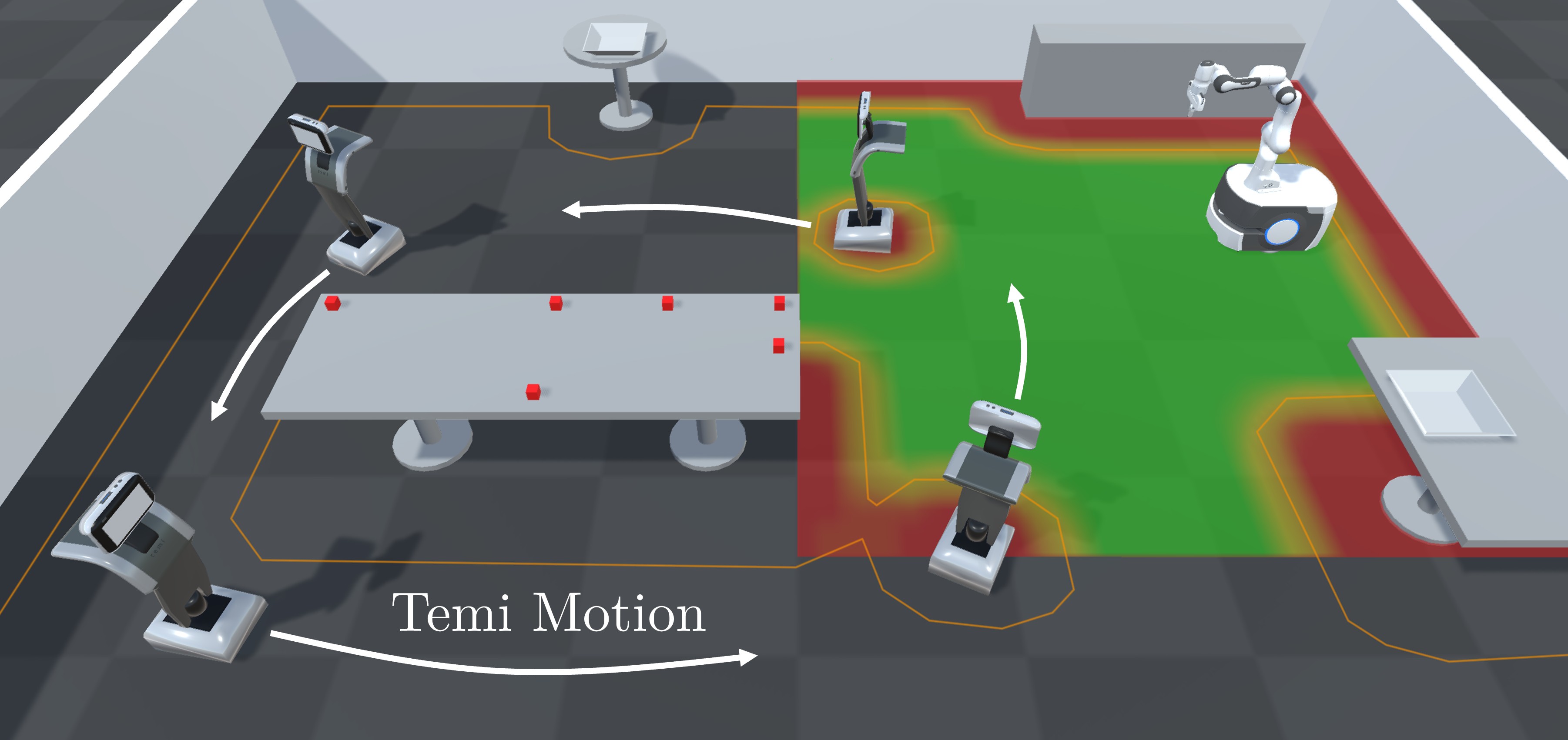}}
\caption{Simulated experiment 2 with four autonomous Temi robots providing dynamic obstacles. }                       
\label{fig:Experiment2}
\vspace{-15pt}
\end{figure}

This experiment uses the same basic arrangement as experiment 1 but introduces additional autonomous robots to the scene which function as dynamic obstacles. In simulation, four Temi robots were tasked with autonomously driving around the table (Fig. \ref{fig:Experiment2}). The large number of robots relative to the size of the environment ensures frequent interactions and therefore avoidance manoeuvres. We also perform a real-world trial with a single Temi robot. The complex and unpredictable interactions between the robots make the task time metric difficult to interpret for these experiments. Instead, we demonstrate that our system can complete the task with dynamic obstacles in the real world and examine the data from an example simulated trial to investigate the performance of the obstacle avoidance functionality.

\section{Results}

\subsection{Experiment 1: Table Clearing}
\subsubsection{Simulated Results}
\begin{figure}[t]
\centerline{\includegraphics[width=\linewidth]{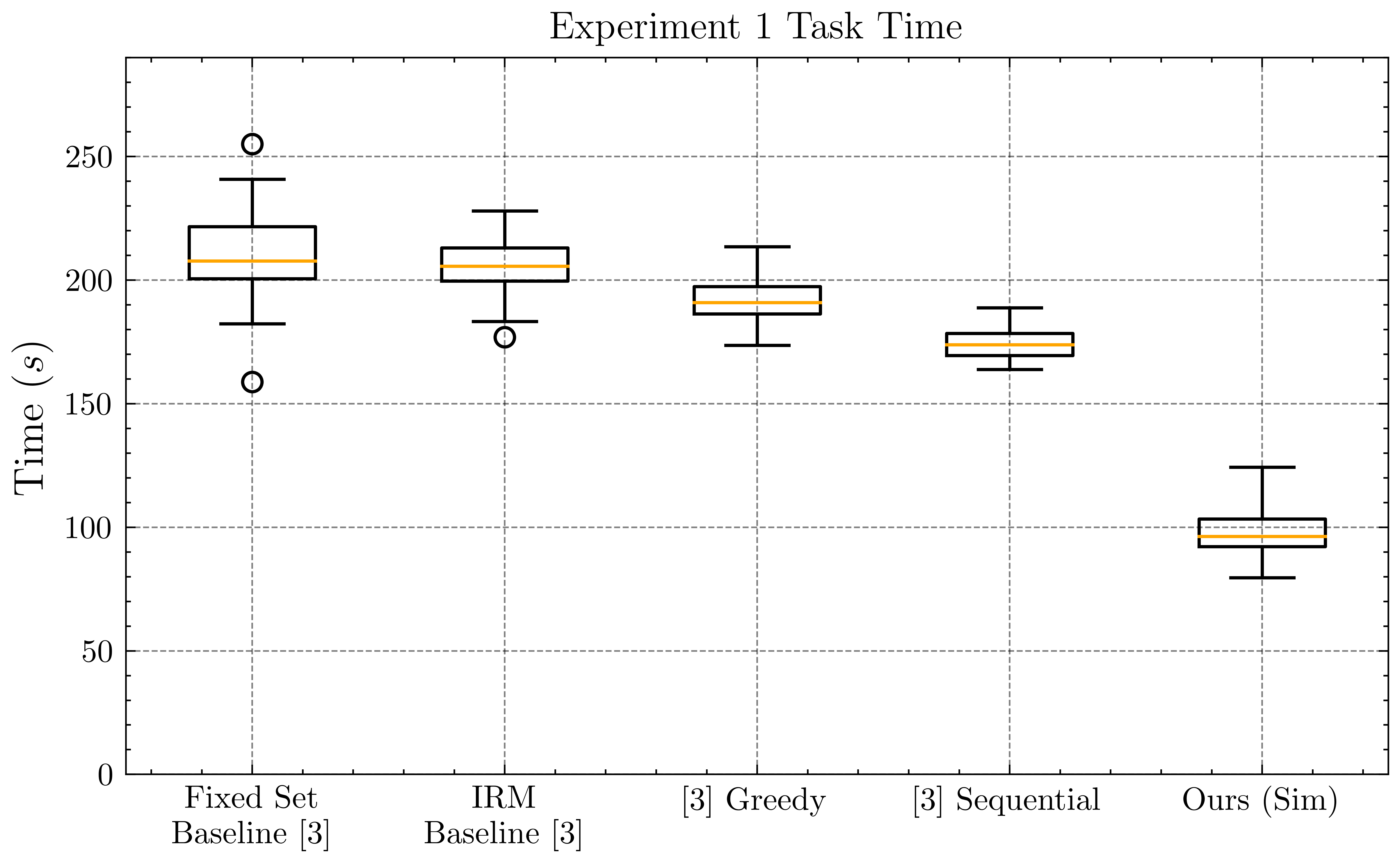}}
\caption{Time required to complete the simulated experiment 1 task. The box plots for the Fixed Set, IRM, Greedy, and Sequential methods have been recreated from the results published in \cite{Reister} which do not include the time required to complete the grasps, which they estimate at about 40\si{s}. Consequently, the timings from \cite{Reister} have been increased by 40\si{s} to compare with our results. For all experiments the objects are rectangular prisms, with comparable grasp complexity.}
\label{fig:Experiment1Results}
\vspace{-10pt}
\end{figure}

The results presented in Fig. \ref{fig:Experiment1Results} compare the time to complete the 6 object pick-and-place task with the baselines and approach presented in \cite{Reister}. Our MotM method reduces the total task time by 43\% compared to the sequential task optimised method presented in \cite{Reister}.

\subsubsection{Real-world Results}
Table \ref{tab:Exp1RealWorldResults} compares the real-world performance of our method with the IRM baseline and Sequential optimised method from \cite{Reister}. On a single trial (6 object pickups) with approximately the same object arrangement and order, we successfully cleared the table in 91 \si{s} where the method presented in \cite{Reister} takes 176 \si{s}, a task time reduction of 48\%. Across 10 trials (60 object pickups) with random object arrangement, the task was completed in an average of 100.4 \si{s} with a success rate of 58 out of 60. The two failure cases resulted from grasp failures where imperfect control of the robot while the fingers were closing caused a collision between the object and fingers that destabilised the grasp. These failures could be recovered by detecting the failure and allowing the robot to reopen the gripper and reattempt the grasp. 

The consistency between the real-world and simulated timing data validates the simulation as a meaningful representation of real-world performance. In general, tasks are completed slightly faster in simulation, which we attribute to idealisations of the perception and low-level controllers. 

\subsection{Experiment 1a: Table Clearing with Static Obstacles}
\begin{figure}[t]
\centerline{\includegraphics[width=\linewidth]{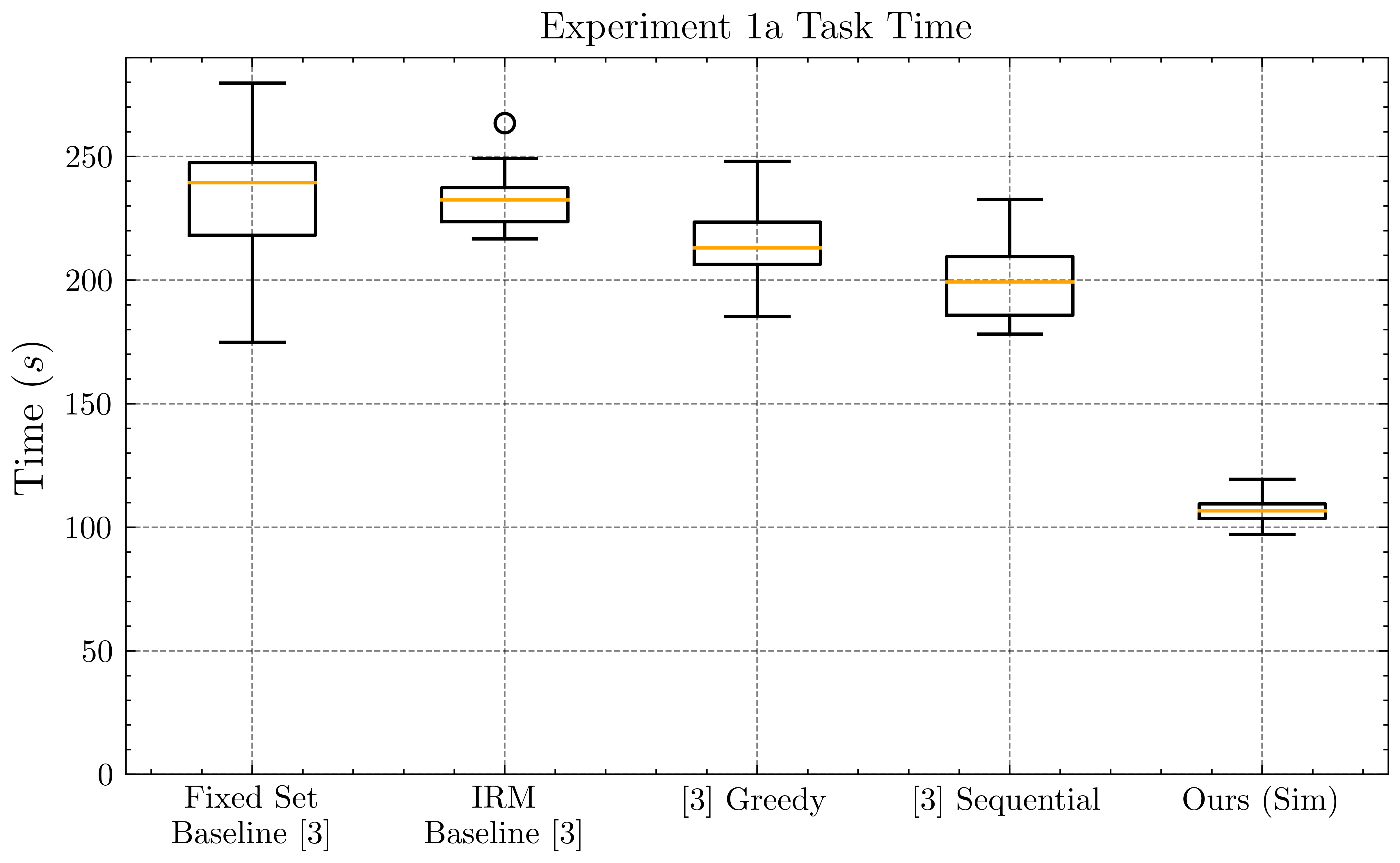}}
\caption{Time required to complete the experiment 1a task with additional obstacles in simulation. Note that as in Fig. \ref{fig:Experiment1Results} results taken from \cite{Reister} include grasping times.}
\label{fig:Experiment1aResults}
\end{figure}

Fig. \ref{fig:Experiment1aResults} presents the simulated task times for the task described in Fig. \ref{fig:Experiment1} with additional static obstacles and compares our method to the baselines and approaches presented in \cite{Reister}. We demonstrate a reduction in total task time of 46\% which is consistent with the performance from experiment 1. These results demonstrate that our approach is robust to the inclusion of additional obstacles in the environment, even when they are close to the grasp locations and interfere with the ideal trajectory for grasping an object on the move.  

\begin{table}[t]
\centering
\renewcommand{\arraystretch}{1.3}
\caption{Comparison of task times for experiment 1 in simulation and the real world. Our real-world results are averaged across 10 different scenarios, where the results for the IRM baseline and Sequential methods are from single trials conducted in \cite{Reister}.}
\label{tab:Exp1RealWorldResults}
\begin{tabular}{cccc}
\hline
\textbf{Method}  & \textbf{Simulation} & \multicolumn{2}{c}{\textbf{Real-World}} \\
 & Task Time & Task Time & Success Rate \\ \hline
IRM Baseline \cite{Reister}    & 192.9 \si{s}     & 202 \si{s} & 6/6 \\
Sequential \cite{Reister} & 172.3 \si{s} & 176 \si{s} & 6/6 \\
Ours (Single Trial) & 85.6 \si{s} & 91 \si{s} & 6/6\\
Ours (Ten Trials)  &  95.9 \si{s} &   100.4 \si{s} & 58/60  \\ \hline
\end{tabular}%
\vspace{-10pt}
\end{table}


\subsection{Experiment 2: Table Clearing with Dynamic Obstacles}
We demonstrate real-world manipulation on-the-move in an environment with unpredictably moving dynamic obstacles. The time required to complete the task is dependent on the random interactions between our robot and the Temi acting as a dynamic obstacle. The complexity of the interactions makes the experiments impossible to properly replicate, and therefore it is meaningless to compare the task-time metric over relatively few trials. However, in an example real-world scenario, our system completed the task in 94 \si{s} in the presence of a dynamic obstacle. In this trial, there were 4 interactions where an evasive manoeuvre was required to avoid the other robot. For the same object arrangement without the dynamic obstacle, the task was completed in 91~\si{s}. This demonstrates that our system can robustly perform tasks in environments with dynamic obstacles while incurring minimal increase in overall task execution time. 

\begin{figure}[t]
\centerline{\includegraphics[width=\linewidth]{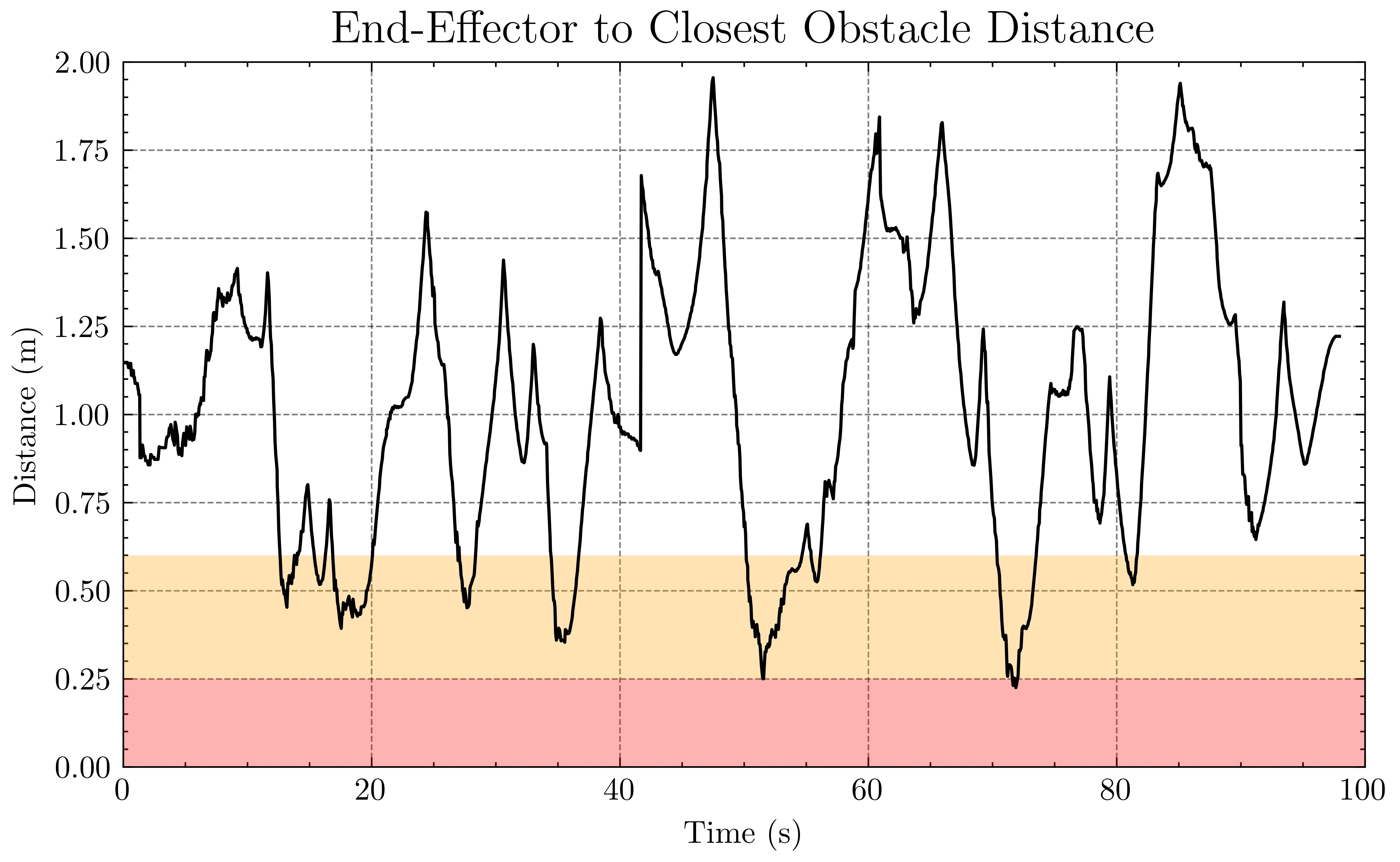}}
\caption{Minimum distance between end-effector and closest detected obstacle over a simulated trial with dynamic obstacles. Discontinuities are caused by obstacles coming into view as the robot moves through the environment. The red and orange regions indicate the minimum end-effector to obstacle distance and region of influence of the constraint described in Section \ref{ArmAvoidance}.}
\label{fig:EndEffectorToObstacle}
\vspace{-5pt}
\end{figure}

Fig. \ref{fig:EndEffectorToObstacle} shows the minimum distance between the end-effector and the closest obstacle over an example simulated trial with 4 dynamic obstacles. These results demonstrate the effectiveness of the arm obstacle avoidance constraints described in Section \ref{ArmAvoidance}. When the distance dips into the region of influence (shown in orange), the constraint limits the velocity toward the obstacle. The minimum distance allowed between the end-effector and obstacles is 0.25 \si{m}. 

\begin{figure}[t]
\centerline{\includegraphics[width=\linewidth]{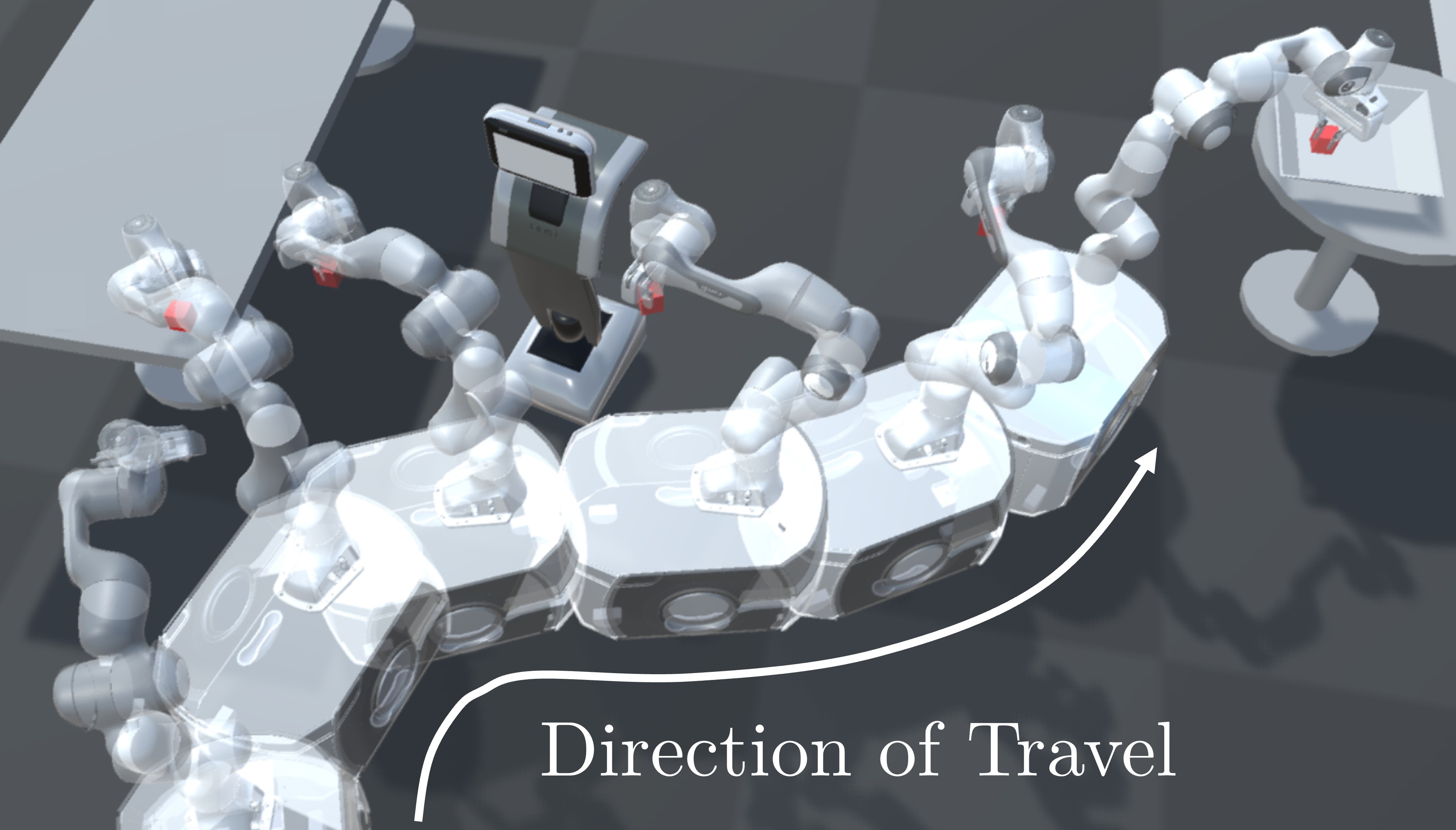}}
\caption{A series of snapshots from our system's arm and base successfully avoiding another robot while performing a pick-and-place task on-the-move.}
\label{fig:TemiAvoid}
\vspace{-20pt}
\end{figure}

Fig. \ref{fig:TemiAvoid} illustrates what the behaviour of the robot looks like in practice. After reaching out to grasp the object, the system retracts the arm to avoid the Temi while driving on towards the drop location to complete the task. 

\section{Discussion and Future Work}
There are numerous avenues that can be explored in future work to improve performance.

\subsubsection{Obstacle Motion Prediction}
\begin{figure}[t]
\centerline{\includegraphics[width=\linewidth]{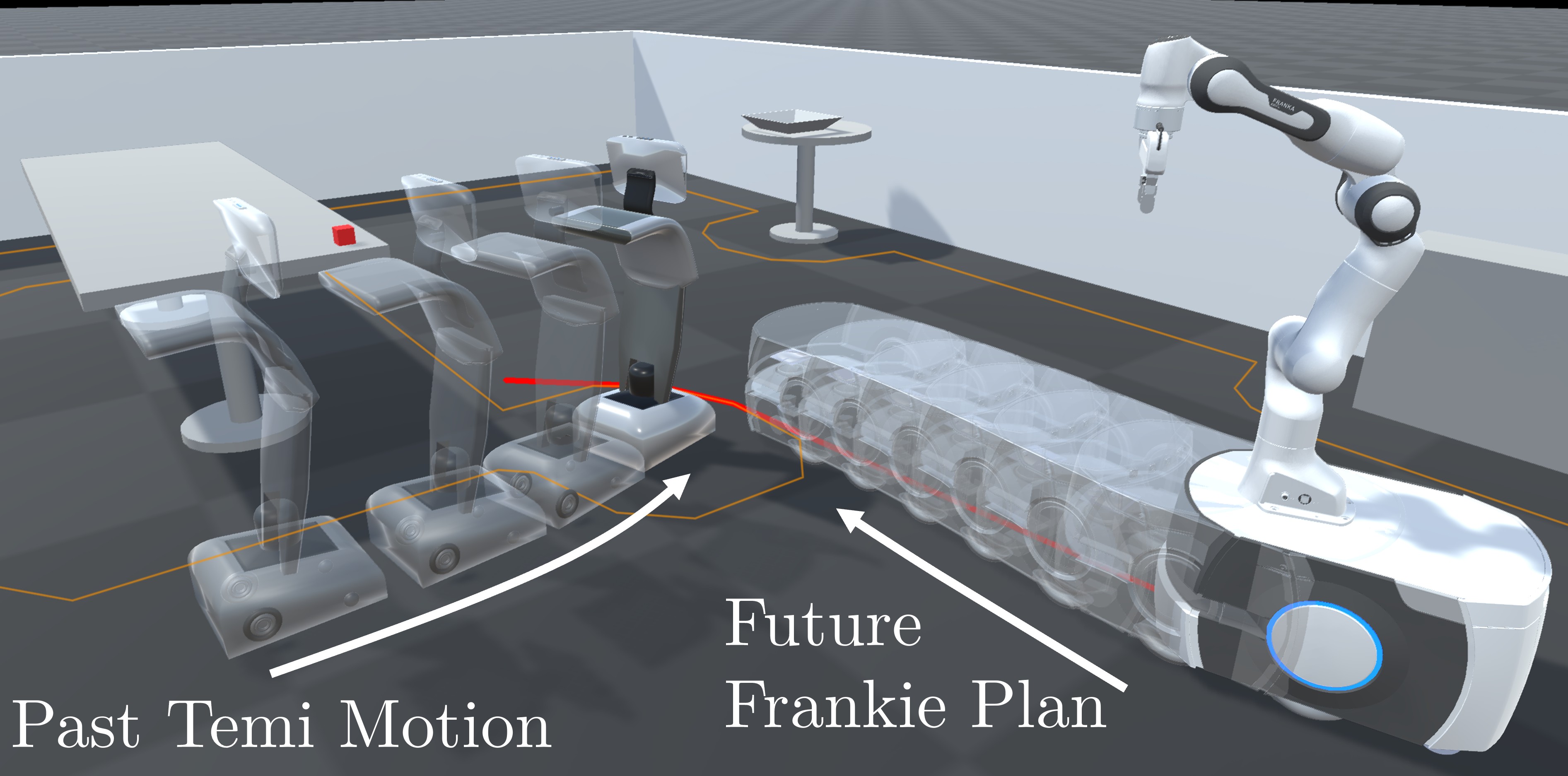}}
\caption{Demonstration of our system's tendency to cut in front of dynamic obstacles. Transparent robots illustrate Temi's history and Frankie's planned path.}
\label{fig:DynamicObstaclePrediction}
\vspace{-20pt}
\end{figure}

Our current implementation has no predictive capability for the motion of dynamic obstacles. Instead, dynamic obstacles are detected at each time step and treated the same as static obstacles for path planning. We have noticed that this results in a tendency for our system to cut in front of other robots and people in the environment. For example, consider the case illustrated in Fig. \ref{fig:DynamicObstaclePrediction}. If we assume that the Temi is stationary then the optimal path to the object (red cube) passes in front of the obstacle. This is the path that is currently taken by our system. However, if the Temi was modelled as a dynamic obstacle, its time history (shown as transparent) could be used to predict that the obstacle will keep driving forward. In that case, a better path is to drive directly toward the object allowing the Temi to move out of the way before our robot arrives. 

The implementation of STAA* presented in \cite{STAA} includes support for dynamic obstacle motion prediction and demonstrates the desired behaviour. However, obstacle motion prediction is not included in this work due to the difficulty of accurately segmenting obstacles from real-world lidar data. A system that explicitly modelled dynamic objects would allow for improved performance in environments with other moving agents. We view this as a perception challenge that could be addressed in future work.  

\subsubsection{3D Collision Geometry for Arm}
As discussed in Section \ref{ArmAvoidance}, collision avoidance for the arm is enabled by assuming that all obstacles detected by the base lidar are tall enough that they should be avoided by the arm. This is both overly conservative for short obstacles and does not prevent collision with obstacles above the plane of the 2D lidar. Including additional sensors and a perception pipeline that accurately models 3D geometry would enable improved manipulator obstacle avoidance and manipulation in cluttered environments such as reaching into a bin or a cupboard. 

\subsubsection{Failure Recovery}
The base control system presented in this work reactively adjusts the base motion to keep the object in reach until the grasp is completed. This is more thoroughly investigated in our related work presented in \cite{StaaFailureRecovery}. However, we have not implemented a robust method for detecting grasp failures in the real world. The robot's gripper provides infrequent feedback on the finger position and applied forces, which makes it difficult to detect failures through proprioceptive methods. Instead, additional tactile or vision sensors could be used for feedback. If grasp failures could be reliably detected, then the presented system would control the robot such that grasps can be reattempted until success is achieved. This would enable recovery from manipulation failures where a second attempt can be easily executed. However, failures such as the object falling to the floor will likely require more sophisticated recovery behaviours including searching for the object. 

\subsubsection{Gripper Speed}
The relatively slow speed of the Franka-Emika Panda gripper requires the system to stabilise the gripper over the object for a significant period of time (approximately 0.8 \si{s}) while the fingers close. This limits the speed at which the robot can be driving past while the grasp is attempted. A gripper that closed more quickly could be used to enable faster grasping on-the-move. 

\subsubsection{Phantom Collisions}
The Franka-Emika Panda manipulator is a cobot with in-built collision detection implemented through joint torque measurement. This is a valuable safety feature, however, accelerations of the mobile base caused by changes in commanded velocity as well as bumps in the terrain can impart torques on the robot joints that are registered as collisions and result in a pause or shutdown of the arm's controller. Decreasing the frequency of these detections by increasing arm compliance warrants further investigation for higher speed performance.

\subsubsection{Grasp Synthesis for Complex Objects}
The simple perception and grasp synthesis method used in this work limits the system to grasping objects of simple geometry and uniform colour. The addition of a more sophisticated grasp synthesis method would enable the grasping of more complex objects. The key challenge in this respect is developing a grasp synthesis system that provides closed-loop feedback throughout the grasping action, all the way until the fingers close. Although not essential, closed-loop feedback throughout the grasping action improves performance in the highly dynamic environment of manipulation on-the-move, and provides reliable performance that is robust to imprecise perception and localisation, and inaccurate robot control \cite{DGBench}. 

\section{Conclusion}
We presented a base control method that integrates with the architecture presented in \cite{MotM} to enable reactive manipulation on-the-move in complex environments with dynamic obstacles. Further, the QP in the redundancy resolution module is augmented to include an inequality constraint that provides collision avoidance for the manipulator. We have explored the system's performance in a number of simulated and real-world experiments. We demonstrated a reduction in task execution time of 48\% compared to a state-of-the-art method on an example real-world pick-and-place task. In addition, we show that the mobile manipulator can perform pick-and-place tasks while its arm and base both avoid dynamic obstacles. Several limitations of the system are explored which highlight interesting opportunities for further research.  

\bibliographystyle{IEEEtran}
\bibliography{IEEEabrv, references.bib}

\end{document}